\documentclass{Interspeech}

\interspeechcameraready

\usepackage{microtype}
\usepackage{cite} 
\usepackage{url, float, tipa, xstring, amsmath, amssymb, graphicx, xcolor, multicol, multirow}
\usepackage{threeparttable} 
\usepackage{subcaption} 
\usepackage[toc,page]{appendix}

\newcommand{\crv}[2]{{#1}\textbackslash{#2}}
\newcommand{\wv}{w2v2}

\renewcommand{\paragraph}[1]{\vspace{3pt}\noindent\textbf{#1}\ \ }

\title{Analyzing the relationships between pretraining language, phonetic, tonal, and speaker information in self-supervised speech models}

\author[affiliation={1}]{Michele}{Gubian}
\author[affiliation={1}]{Ioana}{Krehan}
\author[affiliation={2}]{Oli}{Liu} 
\author[affiliation={1}]{James}{Kirby}
\author[affiliation={2}]{Sharon}{Goldwater}

\affiliation{Institute for Phonetics and Speech Processing}{Ludwig Maximilian University of Munich}{Germany}
\affiliation{
School of Informatics}{University of Edinburgh}{UK}
\email{m.gubian@phonetik.uni-muenchen.de, Ioana.Krehan@campus.lmu.de, Oli.Liu@ed.ac.uk, j.kirby@phonetik.uni-muenchen.de, sgwater@inf.ed.ac.uk}

\begin{document}

\maketitle

\begin{abstract} 

Analyses of self-supervised speech models have begun to reveal where and how they represent different types of information. However, almost all analyses have focused on English. Here, we examine how wav2vec2 models trained on four different languages encode both language-matched and non-matched speech. We use probing classifiers and geometric analyses to examine how phones, lexical tones, and speaker information are represented. We show that for all pretraining and test languages, the subspaces encoding phones, tones, and speakers are largely orthogonal, and that layerwise patterns of 
probing accuracy are similar, with a relatively small advantage for matched-language phone and tone (but not speaker) probes in the later layers.  Our findings suggest that the structure of representations learned by wav2vec2 is largely independent of the speech material used during pretraining.

\end{abstract}

\section{Introduction}
The success of self-supervised speech models has spurred interest in understanding the nature of the representations that these models learn during pretraining (for an overview, see Section V of \cite{mohamed2022self}).
Previous researchers have addressed questions such as what types of information are available in different layers \cite{pasad_layer-wise_2021,shen_wave_2023,shen2024encoding,de2024layer,yang_what_2023,chowdhury_what_2024,pasad_comparative_2023,pasad_what_2024} and how that information is structured \cite{liu.tang.ea:self-supervised,mohamed.liu.ea:orthogonality}.
However, most work in this area has studied representations from English pretrained models, analyzed using English data. 
As a result, it is unclear to what extent these findings generalize to models pretrained on other languages, or when the pretraining and analysis languages don't match---simulating the situation at the start of a cross-lingual fine-tuning scenario.\footnote{Fine-tuning is now more common starting from multilingual pretrained models. While we do not explore multilingual models here, our findings provide a good basis for future comparisons to those models, particularly since popular ones such as XLSR \cite{conneau_unsupervised_2020} and MMS \cite{pratap_scaling_2023} are based on the wav2vec 2.0 architecture studied here.}

In this study, we analyze wav2vec 2.0 (henceforth, \wv) models \cite{baevski2020wav2vec} pretrained on four different languages (English, French, Mandarin, and Vietnamese), and tested using the same four languages plus Thai. 
We analyze each model's representations of both matched-language data (e.g., we analyze the model pretrained on Mandarin using Mandarin test data) and cross-language data (e.g., we analyze the Mandarin model using test data from a different language).
We use two methods:

\textbf{Layerwise probing analysis.} Inspired by previous layerwise probing studies \cite{pasad_layer-wise_2021,shen_wave_2023,shen2024encoding,de2024layer,yang_what_2023,chowdhury_what_2024,pasad_comparative_2023,pasad_what_2024}, we train three types of linear classifiers (for phones, lexical tones, and speakers) and investigate the probing accuracy at different layers of each model. 
Our layerwise analysis of tones largely recapitulates those of \cite{shen_wave_2023,de2024layer}, but the analyses of speakers and phones extend previous work. To our knowledge, previous layerwise analyses of speaker information in self-supervised models have only explored models pre-trained on English
 \cite{mohamed.liu.ea:orthogonality,chowdhury_what_2024}, and cross-language phone probing results are limited to a single language pair \cite{rodriguez_self-supervised_2024} or to testing only on English \cite{millet_toward_2022}. Our probing results systematically cover a wider range of conditions (pretraining language, test language, and probe type), permitting more general conclusions.
In particular, we find that:

\begin{itemize}[leftmargin=3mm]
\item For all models and languages tested, phone and tone probing results follow the same general pattern seen in earlier studies on English phones: 
probing accuracy for these linguistic categories increases in the early layers of the model, peaks in the middle layers, and decreases again in the late layers. 
\item A difference in performance between cross-language and matched-language phone/tone probes only emerges in the middle layers, suggesting that this is where language-specific representations emerge. 
\item Nevertheless, this difference is relatively small in most cases, which is somewhat surprising since these languages have quite different phonetic and phonotactic characteristics. However, it does explain why cross-language fine-tuning can be effective.

\end{itemize}

\textbf{Geometric analysis.} This analysis is based on \cite{liu.tang.ea:self-supervised,mohamed.liu.ea:orthogonality}, who showed that speaker and phonetic information are captured in nearly orthogonal subspaces in various self-supervised models (all pretrained on English). They hypothesized \cite{liu.tang.ea:self-supervised} that the orthogonality arises from the statistical independence of these two types of information. If that is so, then other independently varying properties of speech, such as lexical tones,
should also be orthogonally encoded.\footnote{Lexical tones are largely expressed through 
pitch modulations, but must be approximately independent of an individual {\em speaker}'s canonical pitch range, or else they would not be able to express speaker-independent information. The typical linguistic analysis of tones as a separate ``tier'' of information \cite{goldsmith1976autosegmental} also implies that tones and {\em phones} are relatively independent, as we discuss further in Section \ref{sec:crv}.}
To test this hypothesis, we use \cite{mohamed.liu.ea:orthogonality}'s 
orthogonality measure to analyze the geometric structure of the tone subspace, and its relationship to the phone and speaker subspaces,
across all our pretraining and analysis languages---again, providing a much broader analysis than previous work.

This analysis reveals that:

\begin{table*}[t]
\begin{threeparttable}[b]
\caption{Data and models used in our study. All corpora are read speech, except the pretraining data for the Vietnamese model, which also contains YouTube data. Evaluation speakers, phones and tones indicate the number of classes that were selected for the analysis. 
}
\centering
\begin{tabular}{lllllcllccc}
 \toprule
{\bf Language} & {\bf Tonal?} & \multicolumn{3}{c}{\bf Pretraining} & & \multicolumn{5}{c}{\bf Evaluation} \vspace{2pt} \\ 
\cline{3-5} \cline{7-11} \noalign{\vskip 2pt} 
 & & Model & Data (hrs) & Spkrs && Data source & Aligner & Spkrs & Phones & Tones \\ 
 \midrule
English & No & \cite{baevski2020wav2vec}\tnote{a} & 960 & 2120
  && LibriSpeech dev-clean \cite{panayotov2015librispeech}
  & MAUS & 40 & 45 & - \\
French & No & \cite{parcollet2024lebenchmark}\tnote{a} & 7,600 & 2338+
   && Vibravox \cite{jhauret-et-al-2024-vibravox}\tnote{b}
  & MFA & 40 & 39 & - \\
Mandarin & Yes & \cite{lu2023context}\tnote{a} & 1,000 & 1991
  && THCHS-30 \cite{wang2015thchs}
  & MFA & 40 & 51 & 4 \\
Vietnamese & Yes & \cite{nguyen2021vietnamese}\tnote{a} & 13,000 & ?
  && VIVOS \cite{luong2016non}
  & MAUS & 40 & 46 & 6 \\
Thai & Yes & - & - & -
  && Global TIMIT Thai \cite{AB2/JY8T3N_2023}
  & MAUS & 40 & 38 & 5 \\
 \bottomrule
\end{tabular}
\label{tab:corpora}
\begin{footnotesize}
\begin{tablenotes}
\item [a] All pre-trained models are from \url{https://huggingface.co}. We used \texttt{wav2vec2-base} (English); \texttt{wav2vec2-FR-7K-base} (French); \texttt{mandarin-wav2vec2} (Mandarin); and \texttt{wav2vec2-base-vi} (Vietnamese).
       \item [b] \url{https://huggingface.co/datasets/Cnam-LMSSC/vibravox}; \emph{speech\_clean, headset\_microphone} subset
     \end{tablenotes}
     \end{footnotesize}
\end{threeparttable}
\end{table*}

\begin{itemize}[leftmargin=3mm]
\item The earlier finding of orthogonality between speaker and phone subspaces \cite{liu.tang.ea:self-supervised,mohamed.liu.ea:orthogonality} extends to models trained on other languages, as well as to the cross-language testing scenario.
\item The other pairs of subspaces (speakers and tones, phones and tones) are also largely orthogonal, as hypothesized, though to a slightly lesser degree than the speaker and phone subspaces. 
\item In most cases, there are no clear differences or patterns in the degree of orthogonality depending on either the pretraining or testing language. An exception is the phone and tone subspaces, whose degree of orthogonality 
varies over the three tonal test languages, correlating with the degree of statistical (in)dependence between phones and tones in each language.
\end{itemize}

In sum, we find that language-specific encoding of linguistic information in \wv\ models is mild, but consistently arises in the middle layers, while speaker information is not encoded consistently in different models. We also add further support to the hypothesis that these models implicitly disentangle statistically independent sources of information into orthogonal subspaces.

\section{Related work}\label{sec:background}

Despite the many studies analyzing self-supervised speech representations (e.g., \cite{pasad_layer-wise_2021,shen_wave_2023,shen2024encoding,de2024layer,yang_what_2023,chowdhury_what_2024,pasad_comparative_2023,pasad_what_2024,martin_probing_2023,choi_self-supervised_2024,wells_phonetic_2022}), only a few have examined non-English models or data. Some of these take different approaches to ours: e.g., comparing  models against human perceptual data \cite{millet_self-supervised_2022} or using visualization 
\cite{linke_what_2023}. 
Of those studies that use probing classifiers, we know only one that considers speaker information, which tested on several languages but used only a (non-\wv) English pre-trained model \cite{chowdhury_what_2024}. 
A few probing studies have examined non-English models, but all in a limited range of scenarios. Specifically, \cite{de_seyssel_probing_2022} analyzed LSTM models trained on French and English, but only in the matched-language condition;  \cite{rodriguez_self-supervised_2024} tested \wv\ models on both matched- and non-matched conditions, but only for a single language pair (Hindi-English); and \cite[in Supplementary Information]{millet_toward_2022}  reported results for \wv\  models pretrained on Dutch, Mandarin, and French, but only tested on English.  
There are also phonetic probing results for multilingual pre-trained models, but with no detailed analysis by language \cite{baevski_unsupervised_2021,xue_sshr_2024}.
Finally, two studies probed
the representation of lexical tone in \wv\ models trained and analyzed on tonal and non-tonal languages \cite{shen2024encoding,de2024layer}. Some of our probing analyses replicate parts of \cite{shen2024encoding,millet_toward_2022}, but overall we report a broader array of conditions than past work.

As for geometric analyses, these are well-established for text models (e.g., \cite{mimno_strange_2017,cai_isotropy_2020,hernandez_low-dimensional_2021,chang_geometry_2022,park_linear_2023}), but much less so for speech models. We know of only a few examples, most of which look at other types of models (e.g., acoustic word embeddings \cite{abdullah_analyzing_2022} or end-to-end ASR systems \cite{stephenson_untangling_2019}). We build here on the work of \cite{liu.tang.ea:self-supervised,mohamed.liu.ea:orthogonality}, described above, who (like us) examined self-supervised models, but whose studies were limited to English.

\section{Data and models}\label{sec:methods}

The materials used in our study are  summarized in Table~\ref{tab:corpora}.
We use 12-layer \wv\ models \cite{baevski2020wav2vec} pretrained on two tonal languages (Mandarin, Vietnamese) and two non-tonal languages (English, French), plus an equivalent untrained (randomly initialized) model as a baseline. We chose these models to maximize comparability both between the models---which all have the same architecture and number of layers---and with the previous tone probing results of \cite{shen2024encoding}, who used the same models.

Our analyses use evaluation data from the same four languages, plus Thai, for which we could not obtain a pretrained model. In all cases, the evaluation data are distinct from the pretraining data. Both the probing and orthogonality analyses require each speech frame to be labeled with a speaker, tone, or phone. For speakers, we use the corpus metadata. For tones and phones, we used either MAUS \cite{kisler2017multilingual} or MFA \cite{mcauliffe17_interspeech} to obtain forced alignments. We excluded non-speech sounds and a handful of rare phones that did not occur with all speakers.
\tablename~\ref{tab:corpora} lists the number of phone and tone classes per language.\footnote{\label{fn:tone-labels}
Mandarin contrasts four lexical tones plus a fifth `neutral' tone which mainly occurs in unstressed syllables \cite{cao1992neutral}. 
Following \cite{shen2024encoding}, we disregard the neutral tone and consider only the lexical tone of each syllable, i.e. we do not account for tone sandhi. 
Vietnamese is sometimes analyzed as having eight tones, two of which are restricted to syllables with obstruent codas, but only six tones are distinguished in the orthography.
Neither Thai nor Vietnamese have tone sandhi of the type found in Mandarin. The three languages are genetically unrelated.
}

For each model and each test corpus, we extracted representations from all the transformer layers (numbered from 1 to 12) as well as from the output of the convolutional feature extractor (layer 0). All representations are 768-dimensional vectors of real numbers and correspond to a single 20ms input frame of speech. These representations, together with their speaker, tone, and phone labels, were used in all analyses described below.

\begin{figure*}[t]
     \centering
     \begin{subfigure}{0.94\textwidth}
         \centering
         (a) \begin{minipage}[c]{.96\textwidth}
\includegraphics[width=\textwidth]{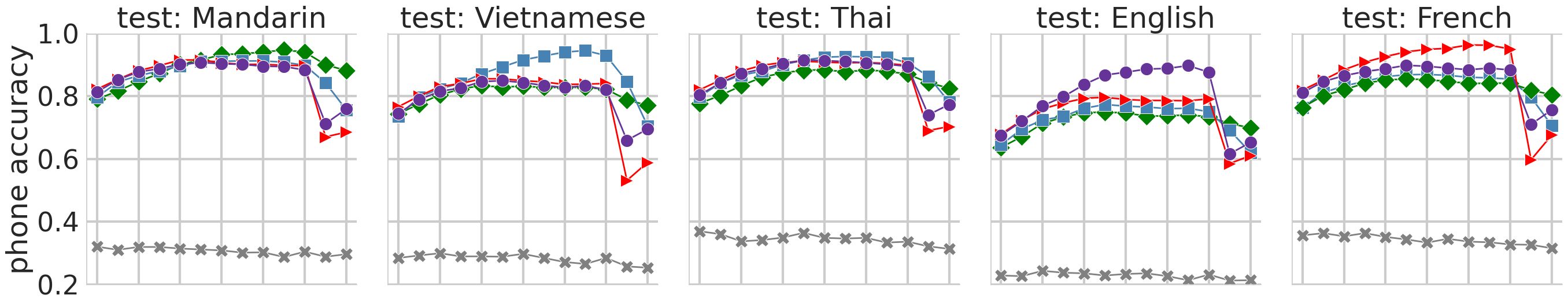}\end{minipage}
\phantomcaption
         \label{fig:acc_phone}
     \end{subfigure}
     
     \begin{subfigure}{0.94\textwidth}
         (b) \begin{minipage}[c]{.96\textwidth}
         \includegraphics[width=\textwidth]{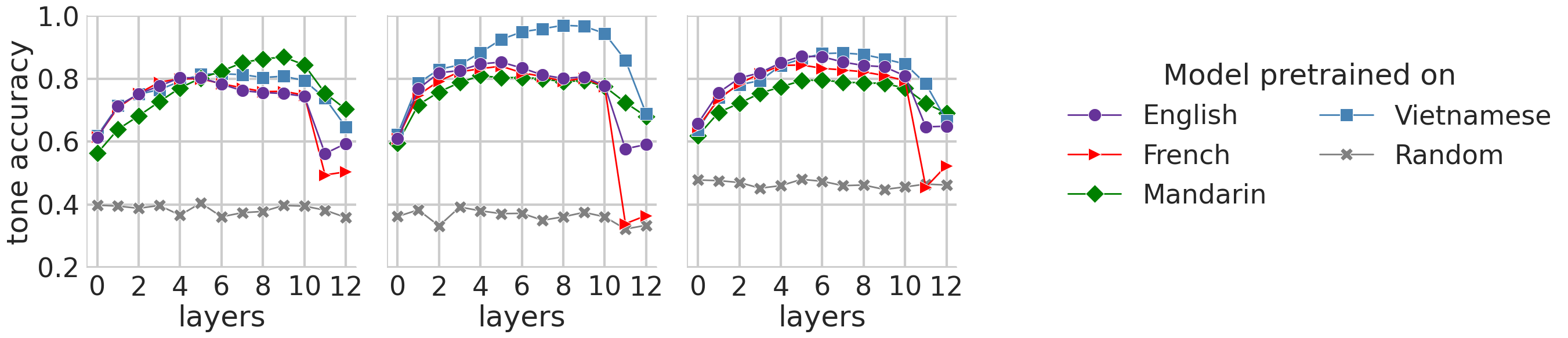}
\end{minipage}
\phantomcaption
         \label{fig:acc_tone}
     \end{subfigure}     

        \caption{Layerwise probing classifier accuracy for (a) phones and (b) tones, across five different test languages. Each plot shows results from four different pretrained models and an untrained (random) model on a single test language.}
        \label{fig:acc}
\end{figure*}
\vspace{-2mm}

\begin{figure}[t]
     \centering
         \includegraphics[width=.48\textwidth]{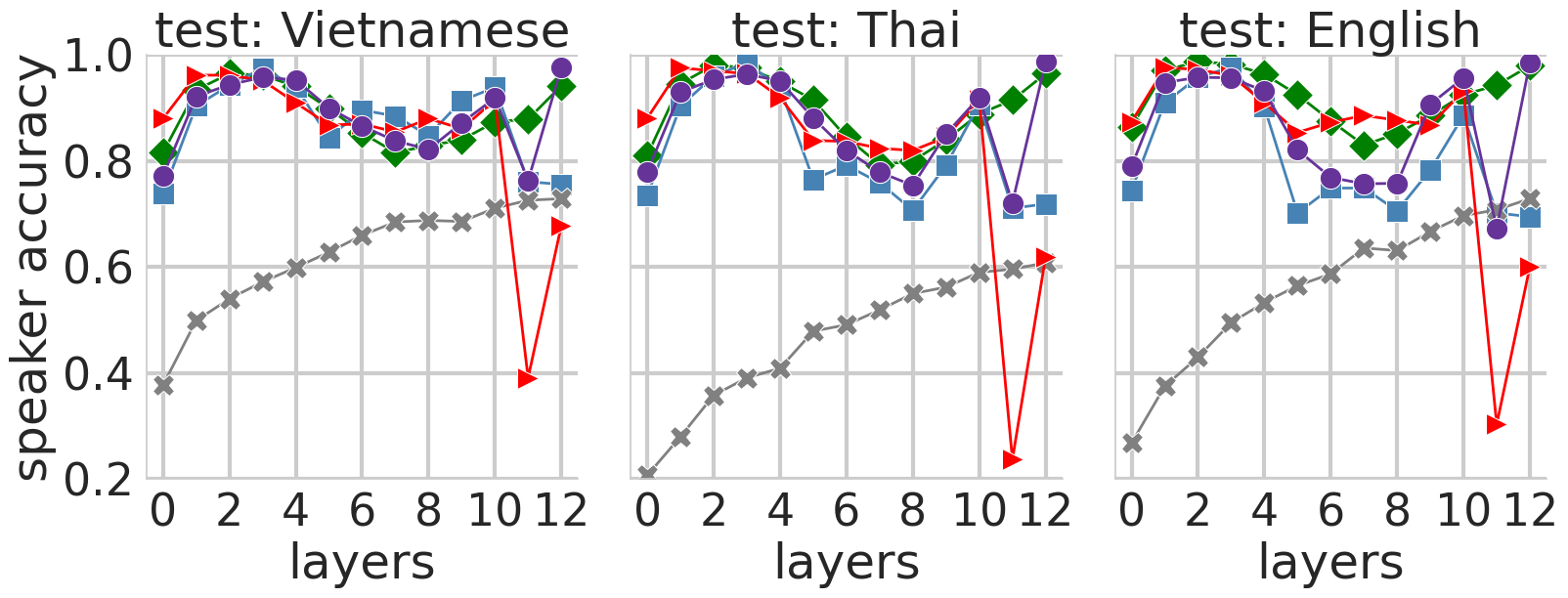}
        \caption{Speaker probing accuracy on three test languages, for the same five models (using the same key) as in Fig.~\ref{fig:acc}. The other two test languages (not shown) display similar patterns. \vspace{-2mm}}
         \label{fig:acc_speaker}
\end{figure}

\begin{figure}[t]
     \centering
     
     \begin{subfigure}{0.48\textwidth}
         \centering
         (a) \begin{minipage}[c]{.94\textwidth}
\includegraphics[width=\textwidth]{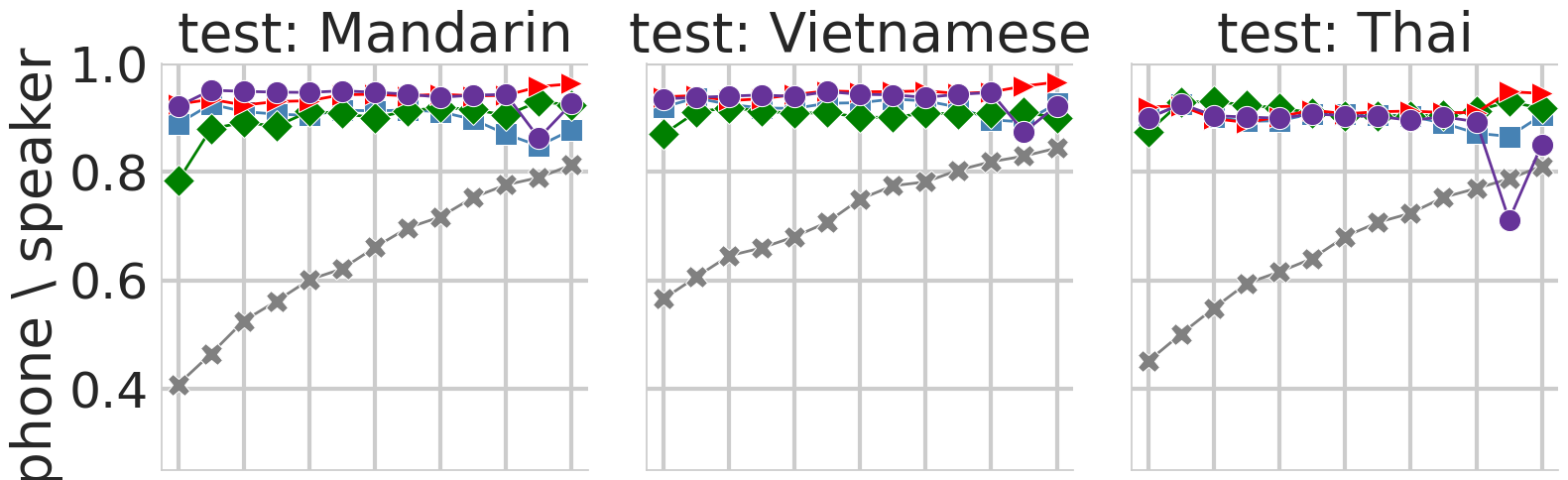}
\end{minipage}
\phantomcaption
         \label{fig:crv_ph_spk}
     \end{subfigure}
    
     \begin{subfigure}{0.48\textwidth}
         \centering
         (b)\hspace{-.1mm} \begin{minipage}[c]{.94\textwidth}
\includegraphics[width=\textwidth]{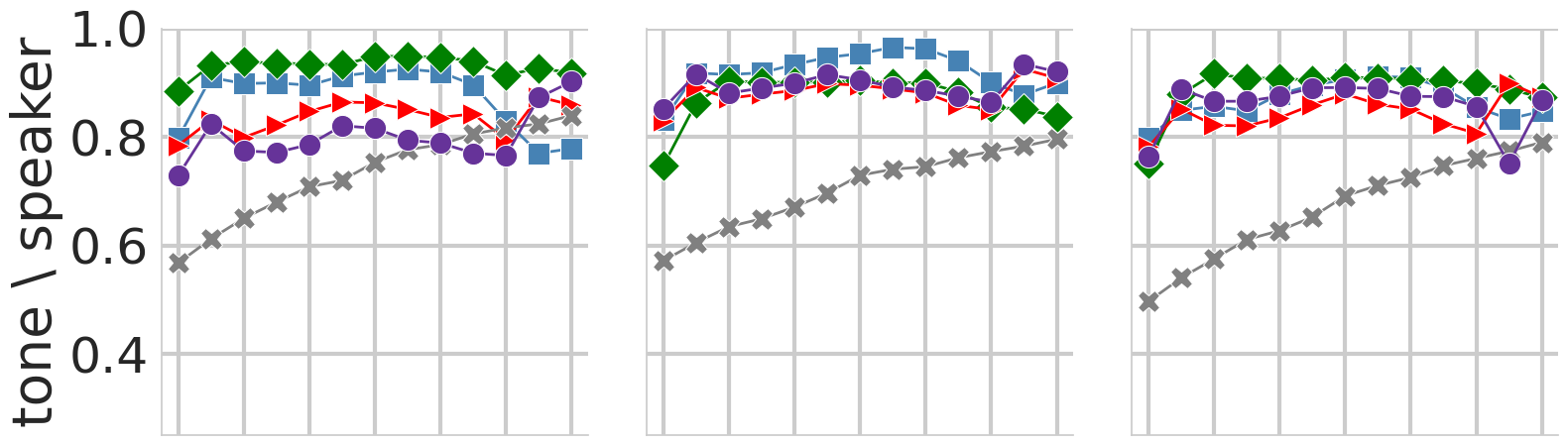}
\end{minipage}
\phantomcaption
         \label{fig:crv_tn_spk}
     \end{subfigure}

    \begin{subfigure}{0.48\textwidth}
         \centering 
         (c) \begin{minipage}[c]{.945\textwidth}
\includegraphics[width=\textwidth]{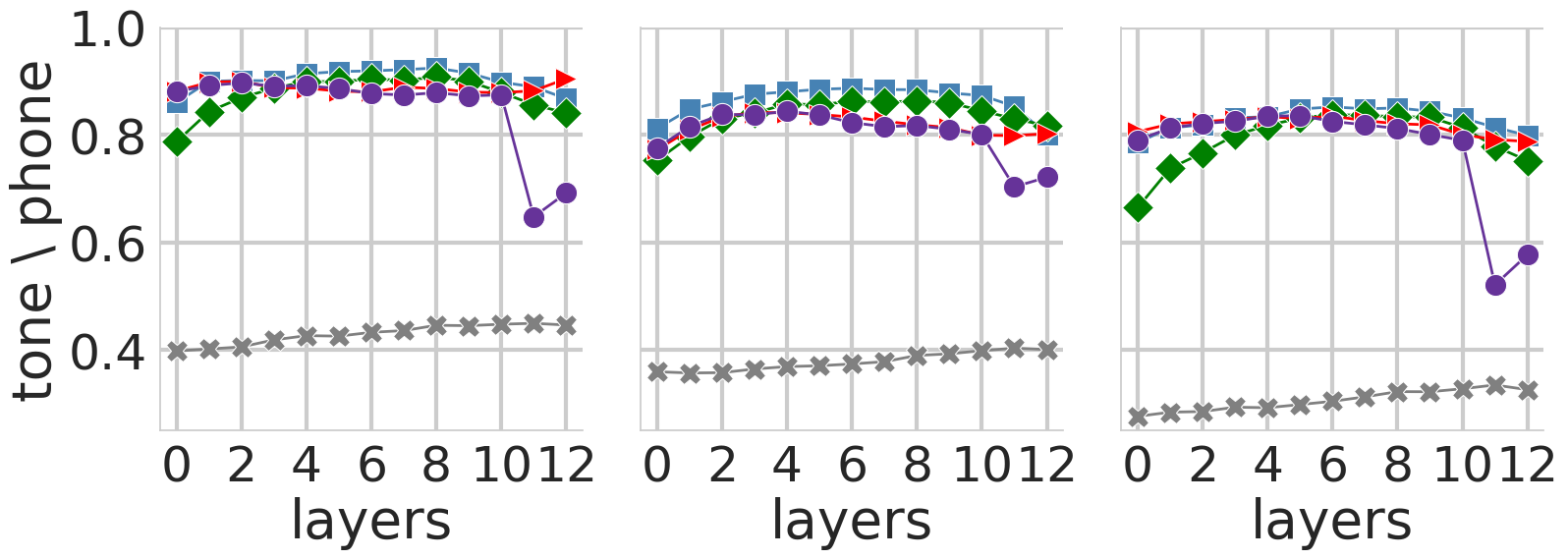}
\end{minipage}
\phantomcaption
         \label{fig:crv_tn_ph}
     \end{subfigure}

\caption{CRV orthogonality measures for (a) \crv{Phone}{Speaker} (only tonal test languages are shown, for space reasons), (b) \crv{Tone}{Speaker}, and (c) \crv{Tone}{Phone}.  Results for the additional CRV measures (\crv{Speaker}{Phone}, \crv{Speaker}{Tone}, and \crv{Phone}{Tone}) can be found in Appendix \ref{appendix_A} and display qualitatively similar patterns. \vspace{-2mm}}
        \label{fig:crv}
\end{figure}

\section{Analysis 1: Probing classifiers}\label{sec:method_probes}

In this analysis, we train linear classifiers to predict phone, tone, or speaker labels from the representations extracted from different models/layers, to quantify the degree to which these types of information are linearly encoded.

As in \cite{shen2024encoding,rodriguez_self-supervised_2024,de_seyssel_probing_2022,martin_probing_2023}, the inputs to the phone and tone classifiers were obtained by averaging the representations along the duration of the same labeled segment (phone or tone).
The inputs to the speaker classifier were created by taking the corresponding phone data sets and substituting phone labels with the appropriate speaker labels. 

Training and test sets for the classifiers were created by randomly sampling (input, label) pairs.
For simplicity, we did not use distinct sets of speakers for the training and test sets, partly because \cite{mohamed.liu.ea:orthogonality} reported that speaker-dependent and speaker-independent phone probing accuracies were very similar and strongly correlated.
The training set size was fixed at about 25k pairs, which was safely beyond training accuracy saturation for all class types.
The test set size was fixed at 10k pairs, which yields a 95\% confidence interval on classification accuracy of about $\pm 1\%$. 
Probing classifiers were linear, fully connected neural networks implementing multinomial logistic regression with cross-entropy loss, using the Adam optimizer with learning rate $10^{-3}$. Training was run for five epochs. Testing was carried out on the classifier state after the last training epoch.

\subsection{Results}\label{sec:acc}

\paragraph{Phones and Tones} 
\figurename~\ref{fig:acc} shows the results of probing classifiers for the two linguistic categories (phones and tones).
In almost all cases, phone probing accuracy is higher than tone probing accuracy, despite the latter having far fewer classes. This indicates that tones are less easily accessible (linearly separable) in the models than phones. Qualitatively, the two categories show a similar pattern: accuracy rises initially, peaks in the middle layers, and drops in the final layers. This is consistent with previous phonetic analyses of \wv\ models for English \cite{pasad_layer-wise_2021,pasad_comparative_2023,yang_what_2023,rodriguez_self-supervised_2024} and Hindi \cite{rodriguez_self-supervised_2024}, and the results of \cite{shen2024encoding,de2024layer} on tones.

We also see a consistent advantage when the pretraining and testing language are matched. While this effect has been found previously for tone probing on several languages \cite{shen2024encoding} and for phone probing on Hindi/English 
\cite{rodriguez_self-supervised_2024}, our results suggest that the effect is very general across both languages and linguistic categories. In addition, our more comprehensive results highlight that the matched-language advantage seems to arise consistently around layers 4-6, with the matched language probe continuing to increase in accuracy while the non-matched probes stabilize or begin to drop. This suggests that language-specific representations only begin to emerge in the middle layers of the models.
Finally, we note that the matched-language advantage is strongest for Vietnamese, possibly due to its much larger and more diverse pretraining data. Further work is needed to fully tease apart the effects of language, data size, and speech style.

\paragraph{Speakers} 
The results for speaker probing (\figurename~\ref{fig:acc_speaker}) are also roughly consistent across models and test languages, but unlike for phones and tones, we see no evidence of a matched-language advantage, and results are a bit noiser overall. The layerwise trend is also different, with the highest accuracy in the early layers (around 2-4) and lower accuracy in layers 5-9 where phone and tone probing display the strongest language-specific advantage. The speaker probes are particularly variable in layer 11, where anomalous behavior has previously been shown in the English model \cite{de2024layer,shen2024encoding,pasad_layer-wise_2021}. We find an even larger anomaly in the French model, which in layers 11-12 is even worse than the random model. (This anomaly is also visible in Figures~\ref{fig:acc} and~\ref{fig:crv}, and is analyzed further in Appendix  \ref{sec:layer11}). 
\section{Analysis 2: Representational geometry}

We follow \cite{liu.tang.ea:self-supervised,mohamed.liu.ea:orthogonality} in examining how  information is represented geometrically. The analysis involves two steps: 
identifying the subspaces corresponding to phones, tones, and speakers; and measuring the orthogonality between each pair of subspaces. 

We identify each subspace as in \cite{liu.tang.ea:self-supervised}. First, we compute the centroid of each class label: e.g., for the phone subspace, we average all embeddings corresponding to each phone class. 
This yields a matrix of size $N_c \times 768$, where $N_c$ is the number of phone classes. By performing PCA on this matrix, we obtain the principal components (PCs) that span the phone subspace. We obtain the PCs of the speaker and tone subspaces analogously.

We measure orthogonality between these subspaces using Cumulative Residual Variance (CRV) \cite{mohamed.liu.ea:orthogonality}.
Consider two different subspaces of $R^n$, defined by the PCs of matrices $X$ and $Y$.
The CRV of $X$ removing $Y$, written \crv{$X$}{$Y$}, measures how much of the variance defined by the subspace $X$ remains after projecting out (``collapsing'') $Y$. CRV accounts for the fact that a given PC of $X$ is aligned to different degrees with a number of directions in $Y$, as well as that being aligned to a PC explaining more (less) variance in $X$ means that $Y$ is more (less) aligned to $X$.
A value of 1 indicates that $X$ and $Y$ are orthogonal (projecting out the PCs of $X$ has no effect on $Y$); lower values indicate greater alignment between the PCs of the two spaces; and values close to 0 are obtained when most of the variance of $X$ is in its first PC, and that PC is also closely aligned to the first PC of $Y$. For a more precise and detailed description of CRV, see  \cite{mohamed.liu.ea:orthogonality}. For the phone and speaker subspaces, we used the top 35 PCs when calculating CRV.

\subsection{Results}\label{sec:crv}

\figurename~\ref{fig:crv} reports CRV measurements. We computed CRV for all combinations \crv{X}{Y}, where X and Y are distinct values of \{Speaker, Phone, Tone\}.  CRVs involving Tone were computed only for the three tone languages.

Looking first at the results involving phones and speakers (Figure~\ref{fig:crv_ph_spk}), we see that the pattern of results previously found on English data using English pretrained (or untrained) models \cite{mohamed.liu.ea:orthogonality} also broadly extends to the other pre-trained models. In particular, phone and speaker subspaces are largely orthogonal, with only small variations across layers (aside from the anomaly in layer 11 of the English model, already mentioned in Section~\ref{sec:acc}). 
Unlike for the probing results, we do not see evidence for any obvious relationship between pretraining language, test language, and degree of orthogonality.

Results involving tones are shown in Figures~\ref{fig:crv_tn_spk}-\ref{fig:crv_tn_ph}. Figure~\ref{fig:crv_tn_spk} shows that projecting out the speaker subspace retains much of the variance in the tone subspace, although the CRV is not as high as for \crv{Phone}{Speaker} (i.e., lower orthogonality), and there is a smaller difference relative to the randomly initialized model. This suggests that the models do not (linearly) disentangle tone and speaker information as much as phone and speaker information. Moreover, here the choice of pretraining language does matter, with the highest \crv{Tone}{Speaker} values observed when the (tonal) pretraining language matches the test language. 

For the \crv{Tone}{Phone} results (Figure~\ref{fig:crv_tn_ph}), we again see relatively high values overall, especially compared to the random baseline. Orthogonality appears to be slightly higher for the models pre-trained on tonal languages, but the largest effect is due to testing language: models tested on Mandarin score higher than those tested on Vietnamese, which score higher than those tested on Thai. If our hypothesis regarding the relationship between statistical independence and orthogonality in the representation space holds, these results suggest that any systematic correlation between tones and phones is highest in Thai and lowest in Mandarin, with Vietnamese falling somewhere in between. 

Although tones are generally analyzed as operating at the syllable level and independent of the segmental structure of the host syllable \cite{goldsmith1976autosegmental}, there is typically some degree of statistical dependence between them in the form of phonotactic constraints, which can range from weak to absolute \cite{kirby2021incorporating}. 
To examine these statistical relationships in the three tone languages of our study, we computed tone-phone co-occurrences for each language using counts extracted from the test corpora. Counts were collected separately for syllable onset, nucleus, and coda, and were then used to  compute the adjusted mutual information \cite{vinh2010information} between tones and phones for each language and syllabic component (\tablename~\ref{tab:MI}). 
The results are consistent with our hypothesis: there is high degree of independence between tones and phones, but also a consistent ordering of Mandarin $<$ Vietnamese $<$ Thai---the same as the ordering of CRV results in \figurename~\ref{fig:crv_tn_ph}. While this is far from conclusive, it suggests that the geometry of the representations may be more dependent on the phonotactics of the embedded language than on the phonotactics of the language material used at pretraining time.

\begin{table}[t]
\caption{Adjusted Mutual Information (AMI) between tones and phones for each language and position in the syllable. AMI ranges from 0 (statistical independence) to 1 (no independence).} 
\label{tab:MI}
\centering
\begin{tabular}{lccc}
\hline
Language & Onset & Nucleus & Coda  \\
\hline

Mandarin & 0.048 & 0.020 & 0.002 \\
Vietnamese & 0.066 & 0.047 & 0.134 \\
Thai & 0.118 & 0.063 & 0.179 \\

\hline
\end{tabular}\vspace{-3mm}
\end{table}

\section{Conclusion}

In this study, we used probing classifiers and geometric analyses to examine how self-supervised \wv\ models represent information about phones, tones, and speakers. Consistent with other studies, accuracy of our phone and tone probes was found to peak in the middle network layers. Our broader analysis of several languages, including cross-language testing, showed that a matched-language probing advantage for both phones and tones emerges consistently in layers 5-9, but is relatively modest.
On the other hand, speaker probing accuracy is somewhat more variable, does not show any matched-language advantage, and peaks consistently in the early layers (and sometimes in the final layer).
Our analysis of the representational geometry suggests that the subspaces encoding phones, tones, and speakers are largely orthogonal, and that orthogonality is minimally affected by choice of pretraining language, possibly reflecting sensitivity to phonotactic constraints. Overall, our findings suggest that the \wv\ architecture learns to encode both linguistic and non-linguistic information in a largely language-independent way, although some language-specificity is observed in the representation of segmental and suprasegmental features in the mid to late layers.

\bibliographystyle{IEEEtran}
\bibliography{refs}

\begin{appendices}

\section{Complete set of figures}\label{appendix_A}

\begin{figure*}[th!]
     \centering
     \begin{subfigure}{0.94\textwidth}
         \centering
         (a) \begin{minipage}[c]{.96\textwidth}
        \includegraphics[width=\textwidth]{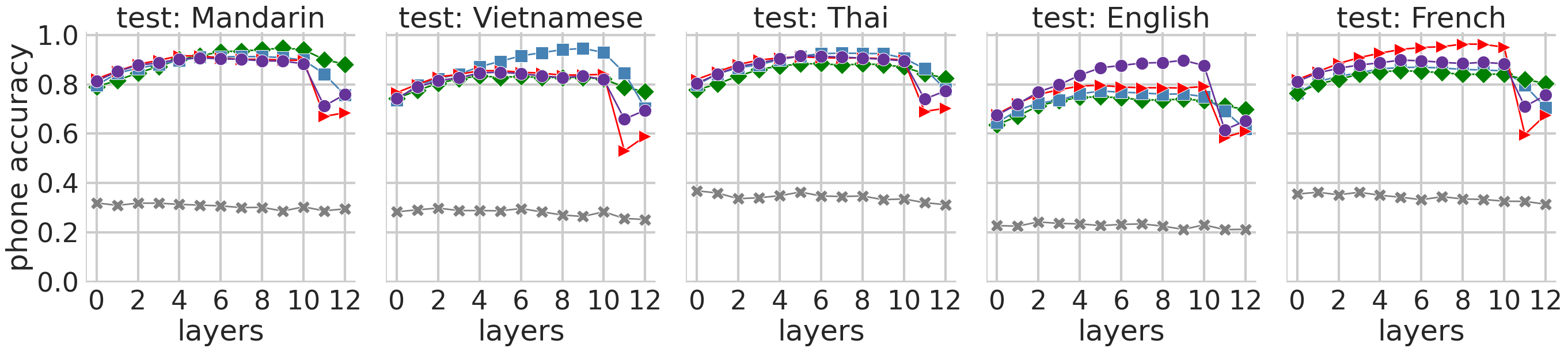}\end{minipage}
\phantomcaption
         \label{fig:acc_phone_apx}
     \end{subfigure}
     
     \begin{subfigure}{0.94\textwidth}
         (b) \begin{minipage}[c]{.96\textwidth}
         \includegraphics[width=\textwidth]{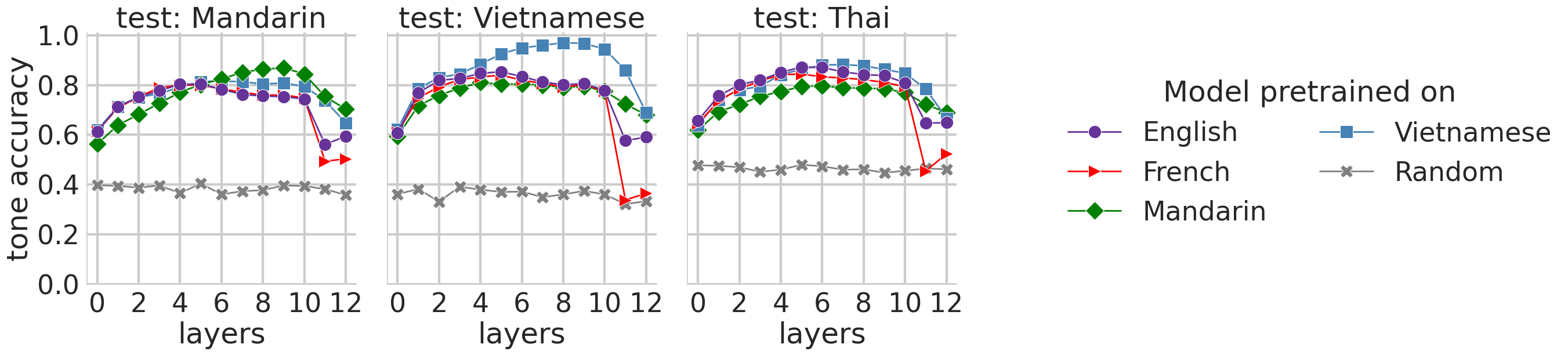}\end{minipage}
\phantomcaption
         \label{fig:acc_tone_apx}
     \end{subfigure}
     
    \begin{subfigure}{0.94\textwidth}
         (c) \begin{minipage}[c]{.96\textwidth}
         \includegraphics[width=\textwidth]{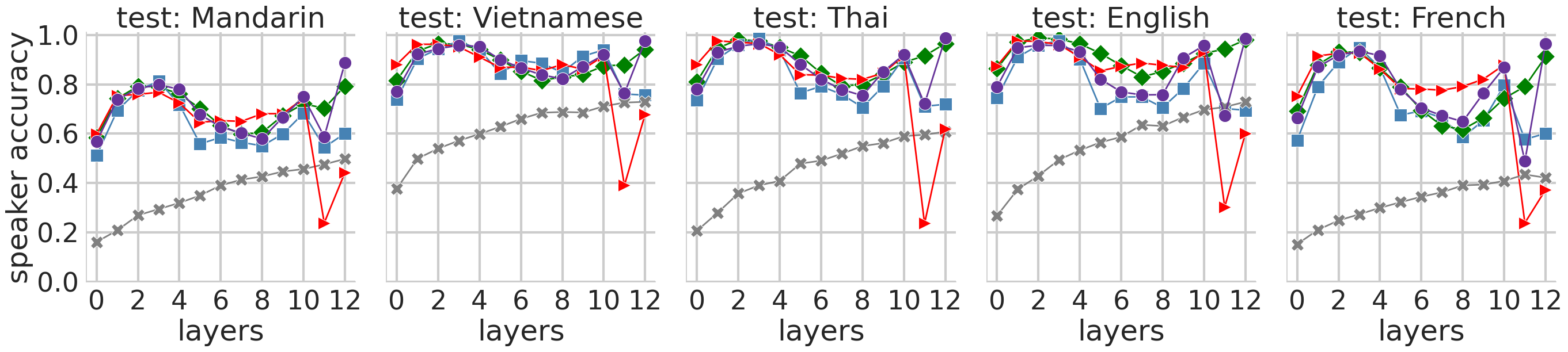}\end{minipage}
\phantomcaption
         \label{fig:acc_speaker_apx}
     \end{subfigure}

        \caption{Layerwise probing classifier accuracy for (a) phones, (b) tones and (c) speakers across five different test languages. Each plot shows results from four different pretrained models and an untrained (random) model on a single test language. {\bf Note}: The overall lower speaker probing accuracy of models tested on Mandarin is probably due to a labeling artifact. The file naming convention in the THCHS-30 corpus suggests that there are 60 speakers, while the reference paper declares 40 speakers (and a confusion matrix of probing results suggests this is closer to the true number). As we based our speaker labeling on the file name coding, there are probably some groups of labels that refer to the same speaker, hence the (apparent) drop in performance. We do not believe this issue would meaningfully affect other results in this paper (notably the orthogonality measures) because splitting a speaker's data would simply create multiple centroids per speaker. As these would be very close to each other, they shouldn't contribute much additional variance to be captured by the PCA on the speaker centroids, and would therefore only affect the least important PCA components.}
        \label{fig:acc_apx}
\end{figure*}

\begin{figure*}[t]
     \centering
     
     \begin{subfigure}{0.94\textwidth}
         \centering
         (a) \begin{minipage}[c]{.96\textwidth}
\includegraphics[width=\textwidth]{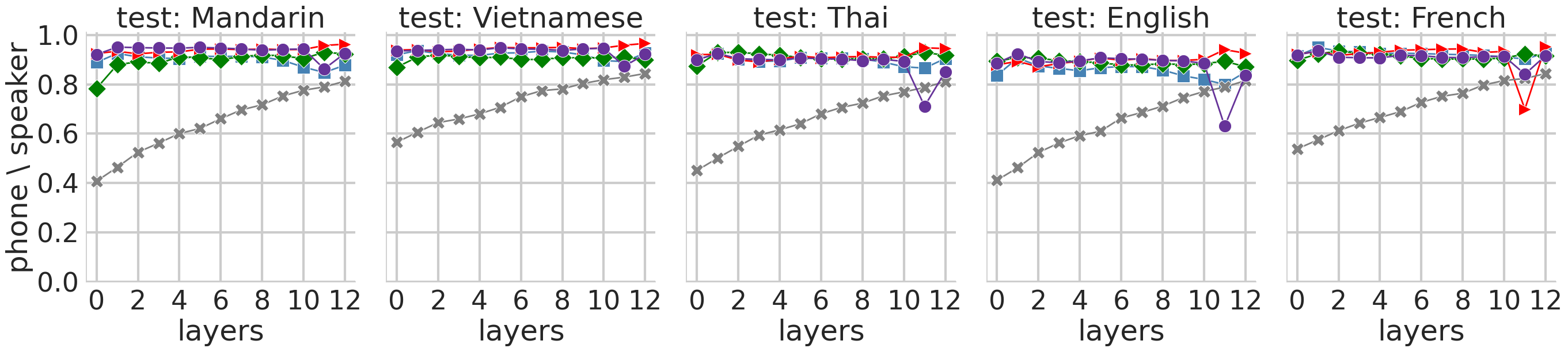}
\end{minipage}
\phantomcaption
         \label{fig:crv_ph_spk_apx}
     \end{subfigure}

     \begin{subfigure}{0.94\textwidth}
         \centering
         (b) \begin{minipage}[c]{.96\textwidth}
\includegraphics[width=\textwidth]{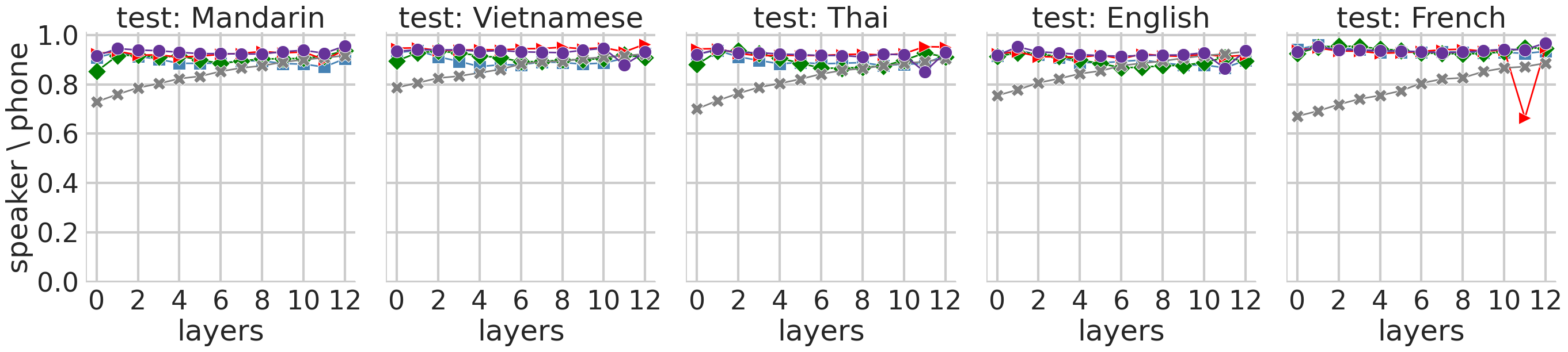}
\end{minipage}
\phantomcaption
         \label{fig:crv_spk_ph_apx}
     \end{subfigure}

     \begin{subfigure}{0.94\textwidth}
         \centering
         (c) \begin{minipage}[c]{.96\textwidth}
\includegraphics[width=\textwidth]{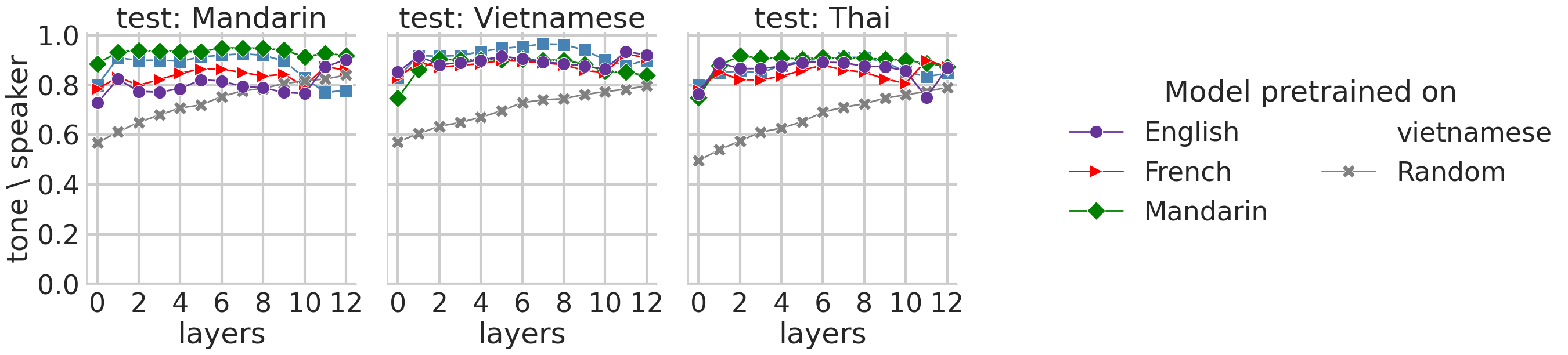}
\end{minipage}
\phantomcaption
         \label{fig:crv_tn_spk_apx}
     \end{subfigure}

     \begin{subfigure}{0.94\textwidth}
         \centering
         (d) \begin{minipage}[c]{.96\textwidth}
\includegraphics[width=\textwidth]{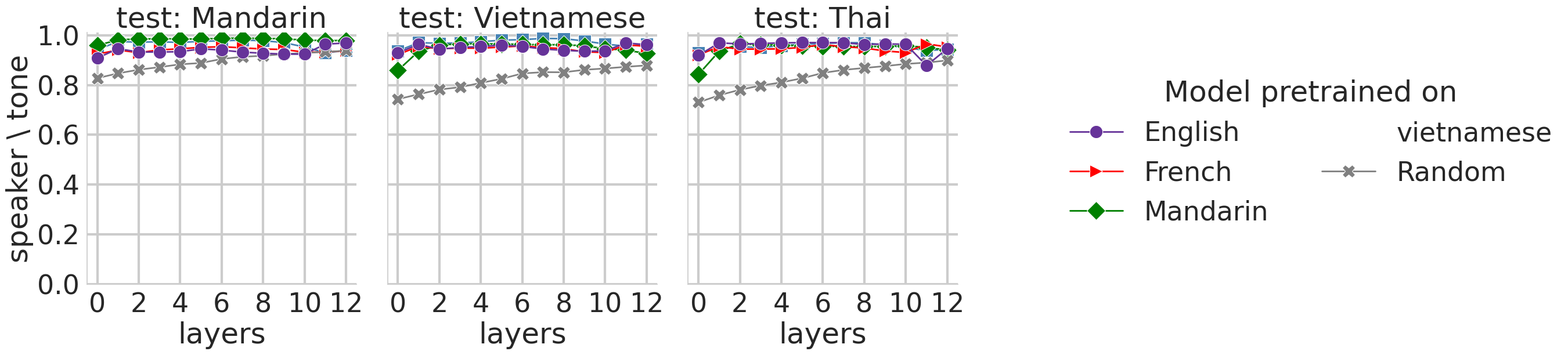}
\end{minipage}
\phantomcaption
         \label{fig:crv_spk_tn_apx}
     \end{subfigure}

    \begin{subfigure}{0.94\textwidth}
         \centering
         (e) \begin{minipage}[c]{.96\textwidth}
\includegraphics[width=\textwidth]{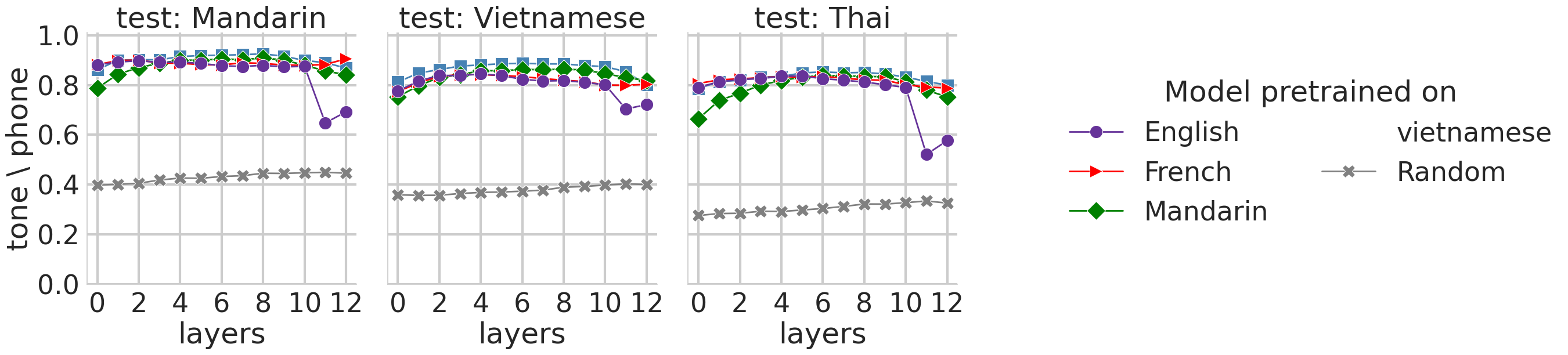}
\end{minipage}
\phantomcaption
         \label{fig:crv_tn_ph_apx}
     \end{subfigure}

     \begin{subfigure}{0.94\textwidth}
         \centering
         (f) \begin{minipage}[c]{.96\textwidth}
\includegraphics[width=\textwidth]{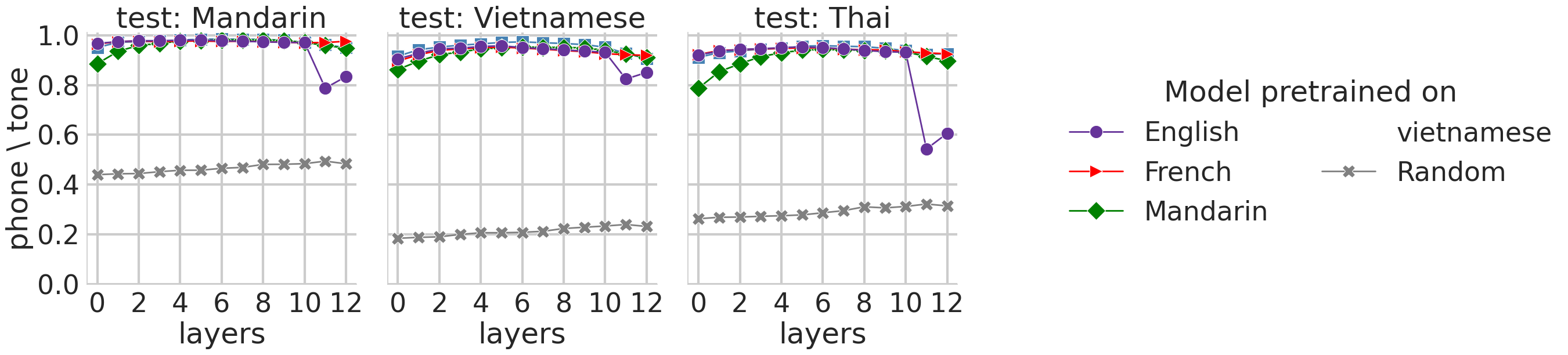}
\end{minipage}
\phantomcaption
         \label{fig:crv_ph_tn_apx}
     \end{subfigure}

\caption{CRV orthogonality measures for
(a) \crv{Phone}{Speaker},
(b) \crv{Speaker}{Phone},
(c) \crv{Tone}{Speaker},
(d) \crv{Speaker}{Tone},
(e) \crv{Tone}{Phone},
(f) \crv{Phone}{Tone}.}
        \label{fig:crv_apx}
\end{figure*}

\figurename~\ref{fig:acc_apx} reports all results of  probing classifiers. \figurename~\ref{fig:crv_apx} reports all CRV measurements.

\section{Anomaly at layer 11}\label{sec:layer11}

\begin{figure*}[h!]
    \centering
    \begin{subfigure}{0.9\textwidth}
        \centering
        (a) \begin{minipage}[c]{.96\textwidth}
\includegraphics[width=\textwidth]{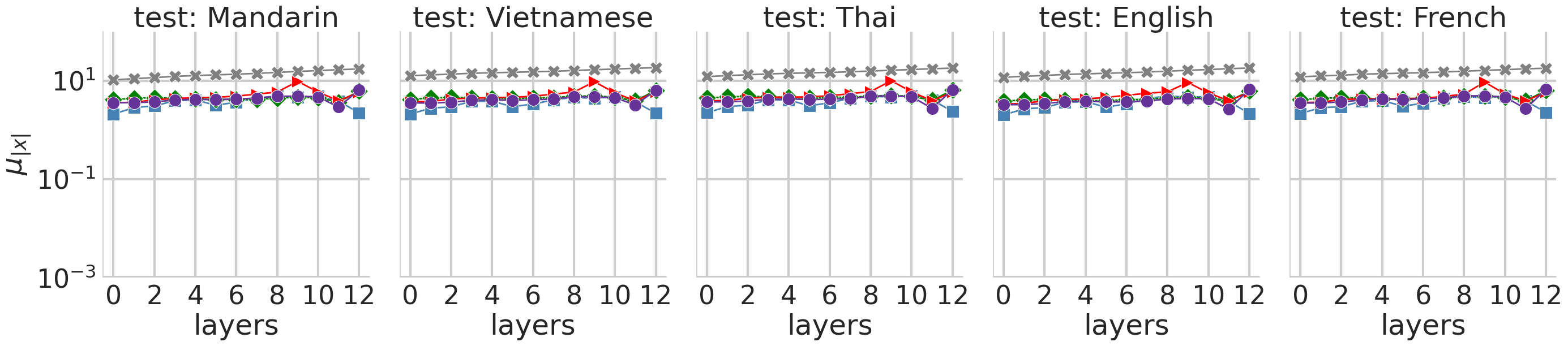}
\end{minipage}
\phantomcaption
        \label{fig:mean_mag_agg_phone}
    \end{subfigure}
    \begin{subfigure}{0.9\textwidth}
        \centering
        \vspace*{3mm}
        (b) \begin{minipage}[c]{.96\textwidth}
\includegraphics[width=\textwidth]{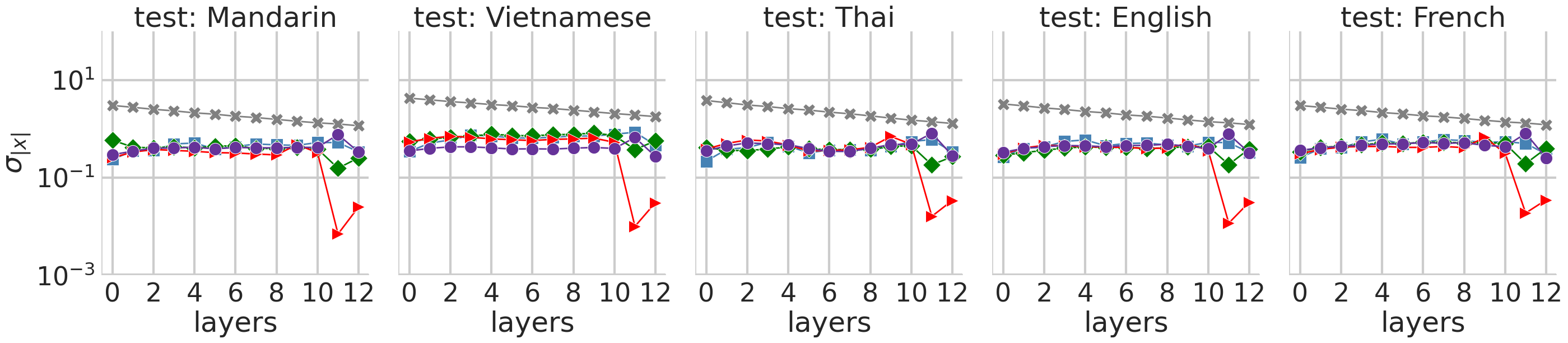}
\end{minipage}
\phantomcaption
        \label{fig:sd_mag_agg_phone}
    \end{subfigure}
    \begin{subfigure}{0.9\textwidth}
        \centering
        \vspace*{3mm}
        (c) \begin{minipage}[c]{.96\textwidth}
\includegraphics[width=\textwidth]{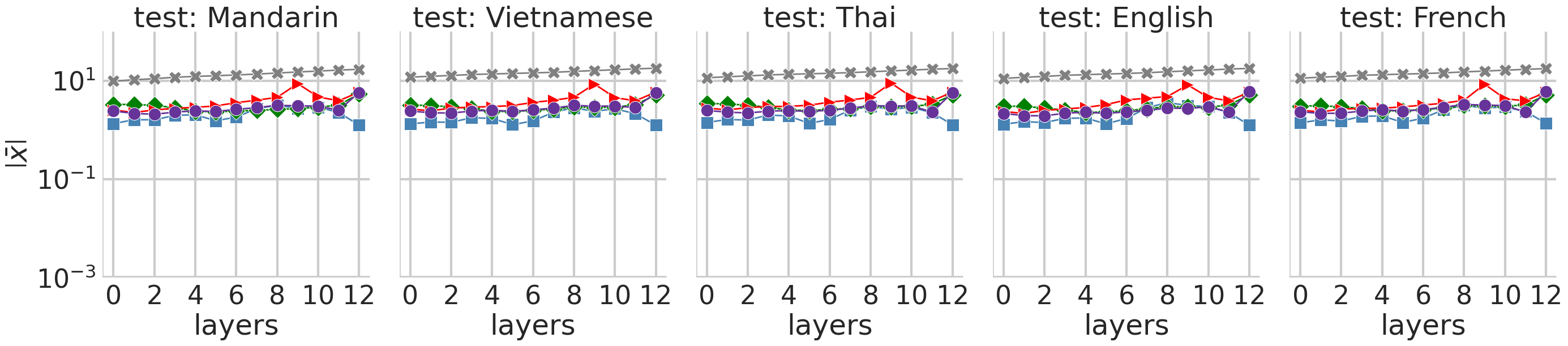}
\end{minipage}
\phantomcaption
        \label{fig:mag_mean_agg_phone}
    \end{subfigure}
        \caption{The quantities defined in Eq.~\eqref{eq:magnitudes} on a log scale: $\mu_{|x|}$ (a), $\sigma_{|x|}$ (b), $\left| \bar{x} \right|$ (c). Color coding is the same as in \figurename~\ref{fig:acc} and \ref{fig:crv}.} 
        \label{fig:layer11}
\end{figure*}

In \figurename~\ref{fig:acc_apx}, a severe drop in probe accuracy values at layer 11 for the representations from the English- and French-trained models are apparent. CRV values in \figurename~\ref{fig:crv_apx} partly exhibit similar trends for the same models. The anomaly for the English checkpoint was also reported by \cite{de2024layer}.
In order to get closer to the origin of such anomaly, we analyzed the magnitude of the representation vectors as follows. For each layer, model and test data, mean and standard deviation of the magnitude of the representation vectors were computed. For simplicity, these computations were conducted on the set of per phone (or per speaker) aggregations, as defined in the main text. Formally, given an aggregate matrix $X$ of dimensions $N_C \times d$, where each row $x_i$ is a $d$-dimensional vector ($d = 768$) representing the mean of a phone (speaker) class, the following three scalar quantities were computed:
\begin{equation}
\begin{aligned} 
\mu_{|x|} = & \frac{1}{N_C} \sum_{i = 1}^{N_C} \left| x_i \right| \\
\sigma_{|x|} = &  \sqrt{\frac{1}{N_C-1} \sum_{i = 1}^{N_C} (\left| x_i \right| - \mu_{|x|})^2} \\
\left| \bar{x} \right| ,\  \bar{x} = & \frac{1}{N_C} \sum_{i = 1}^{N_C} x_i,
\end{aligned}\label{eq:magnitudes}
\end{equation}
where $\mu_{|x|}$ and $\sigma_{|x|}$ are the mean and standard deviation of the magnitudes of the representations,  $\left| \bar{x} \right|$ is the magnitude of the mean of the representations. These statistics computed on the phone aggregate matrices are displayed in \figurename~\ref{fig:layer11}; very similar results were obtained from speaker aggregates. 

The first and the third row of plots report values for $\mu_{|x|}$ and $\left| \bar{x} \right|$, respectively. The fact that these quantities are very similar tells us that at layer 11, as well as in the other layers, representations are more likely to be all in a cloud away from the origin, rather than being on a shell surrounding the origin, which would have resulted in  $\left| \bar{x} \right| \ll \mu_{|x|}$.
A trace of an anomaly appears at layer 11 for the representations from the French-trained model, where the average distance from the origin remains constant but its standard deviation decreases dramatically. This fact may or may not be related to the drop of classification accuracy across the board at layer 11 for those representations visible in \figurename~\ref{fig:acc_apx}. 
A similar phenomenon, though much less pronounced, is also visible for the representations from the Mandarin-trained model, while no major deviation at layer 11 appear for the English-trained model.

While we have no explanation for these anomalies at present, we can at least attest that these are neither English-specific nor strongly dependent on the matching between training and test language. Considering the results reported in Figure 2 of \cite{mohamed.liu.ea:orthogonality}, we are inclined to believe that these anomalies are architecture-specific, as layers 11 and 12 of wav2vec2 stand out in some of the results, compared to all other architectures. 

\end{appendices}
\end{document}